\documentclass[reprint,amsmath,amssymb, aps,pre,floatfix,superscriptaddress]{revtex4-2}

\usepackage{graphicx} 
\usepackage{siunitx}
\usepackage[version=4]{mhchem}
\usepackage{comment}
\usepackage{breqn}
\usepackage{xcolor}
\usepackage{tikz}
\usepackage{scalerel}
\usepackage{xifthen}
\usepackage{chngcntr}
\usepackage{xcolor}
\usepackage{bm}
\usepackage{physics}
\usepackage[utf8]{inputenc}
\usepackage[T1]{fontenc}

\usetikzlibrary{svg.path}
\usepackage[colorlinks,allcolors=blue]{hyperref}

\definecolor{orcidlogocol}{HTML}{A6CE39}
\tikzset{
  orcidlogo/.pic={
    \fill[orcidlogocol] svg{M256,128c0,70.7-57.3,128-128,128C57.3,256,0,198.7,0,128C0,57.3,57.3,0,128,0C198.7,0,256,57.3,256,128z};
    \fill[white] svg{M86.3,186.2H70.9V79.1h15.4v48.4V186.2z}
 svg{M108.9,79.1h41.6c39.6,0,57,28.3,57,53.6c0,27.5-21.5,53.6-56.8,53.6h-41.8V79.1z M124.3,172.4h24.5c34.9,0,42.9-26.5,42.9-39.7c0-21.5-13.7-39.7-43.7-39.7h-23.7V172.4z}
 svg{M88.7,56.8c0,5.5-4.5,10.1-10.1,10.1c-5.6,0-10.1-4.6-10.1-10.1c0-5.6,4.5-10.1,10.1-10.1C84.2,46.7,88.7,51.3,88.7,56.8z};
  }
}

\newcommand\orcid[1]{\href{https://orcid.org/#1}{\mbox{\scalerel*{
\begin{tikzpicture}[yscale=-1,transform shape]
\pic{orcidlogo};
\end{tikzpicture}
}{|}}}}

\begin{document}

\setlength{\unitlength}{1cm}

\def \TITLE {Training microrobots to swim by a large language model}

\def\ADD#1{{\textcolor{magenta}{#1}}}

\def \SI {SI}
\def \viz {\textit{viz.},~}
\def \ie {\textit{i.e.},~}
\def \eg {\textit{e.g.},~}

\def \secone {Viscous Hydrodynamic Environments for Microswimmers}
\def \sectwo {Methodology: LLM Prompt}
\def \secthree {LLM-prompted Swimming Strokes}
\def \secfour {Swimming in a Noisy Environment}
\def \secfive {Prompt Engineering}
\def \seccon {Conclusions and Discussions}

\def \lq {liquid-like}

\def \chain {dynamic chaining}

\newcommand{\figref}[2][{}]{Fig.\
\ref{#2}\ifthenelse{\isempty{#1}}{}{\,(#1)}}

\newcommand{\figrefS}[2][{}]{Fig.~S\ref{#2}\ifthenelse{\isempty{#1}}{}{\,(#1)}} 

\newcommand{\movref}[1]{Movie S{#1}}

\def \mat {Materials and Methods}

\def \jpnas {PNAS}
\def \jnp {Nature Physics}
\def \jnc {Nature Communications}

\def \jprl {Physical Review Letters}

\def \edtnp {Editors of \jnp}
\def \edtnc {Editors of \jnc}

\def \edtpnas {Editorial Board members and Editor of \jpnas}
\def \edtprl {Editors of \jprl}


\def \editor {Editors of Scientific Reports}
\def \journal {Scientific Reports}



\def \dofone {d_1}
\def \doftwo {d_2}
\def \dofc {d_{\text{c}}}
\def \dotdofone {\dot{d_1}}
\def \dotdoftwo {\dot{d_2}}
\def \dotdofone {\dot{d}_1}
\def \dotdoftwo {\dot{d}_2}
\def \dotdofc {\dot{d}_{\text{c}}}

\def \dofmin {d_{\text{min}}}
\def \dofmax {d_{\text{max}}}

\def \brc {\br_{\text{c}}}
\def \bri {\br_{\text{i}}}

\def \brcx {\br_{\text{cx}}}
\def \brcy {\br_{\text{cy}}}
\def \dotbrc {\dot{\br}_{\text{c}}}
\def \dotbrc {\dot{\br}_{\text{c}}}
\def \dotbri {\dot{\br}_{\text{i}}}
\def \u {u}
\def \v {v}
\def \bFi {\bF_i}
\def \bFj {\bF_j}
\def \bFk {\bF_k}

\def \bex {\be_x}
\def \bey {\be_y}
\def \bez {\be_z}

\def \mX {\langle X \rangle}

\def \ns {n_{\text{ht}}}

\newcommand{\eqnrefS}[1]{Eq.~S\eqref{#1}}
\def \De {\text{De}}
\def \Re {\text{Re}}
\def \Ma {\text{Ma}}
\def \Rt {\mR}

\def \mA {\mathcal{A}}
\def \mD {\mathcal{D}}
\def \mC {\mathcal{C}}

\def \be {\mathbf{e}}
\def \bf {\mathbf{f}}
\def \bq {\mathbf{q}}
\def \br {\mathbf{r}}
\def \bt {\mathbf{t}}

\def \bu {\mathbf{u}}
\def \bv {\mathbf{v}}
\def \bx {\mathbf{x}}
\def \bw {\mathbf{w}}

\def \bsigma {\boldsymbol{\sigma}}
\def \btau {\boldsymbol{\tau}}

\def \bzero {\boldsymbol{0}}

\def \bA {\mathbf{A}}
\def \bB {\mathbf{B}}
\def \bC {\mathbf{C}}
\def \bD {\mathbf{D}}
\def \bE {\mathbf{E}}
\def \bF {\mathbf{F}}
\def \bG {\mathbf{G}}
\def \bH {\mathbf{H}}
\def \bI {\mathbf{I}}
\def \bJ {\mathbf{J}}
\def \bK {\mathbf{K}}
\def \bL {\mathbf{L}}
\def \bM {\mathbf{M}}
\def \bN {\mathbf{N}}
\def \bO {\mathbf{O}}
\def \bP {\mathbf{P}}
\def \bQ {\mathbf{Q}}
\def \bR {\mathbf{R}}
\def \bS {\mathbf{S}}
\def \bT {\mathbf{T}}
\def \bU {\mathbf{U}}
\def \bV {\mathbf{V}}

\def \bOmega {\boldsymbol{\Omega}}
\def \bomega {\boldsymbol{\omega}}
\def \bell {\boldsymbol{\ell}}

\def \bGamma {\boldsymbol{\Gamma}}

\def \bn {\mathbf{n}}
\def \bI {\mathbf{I}}

\def \tbu {\tilde{\bu}}
\def \tbr {\tilde{\br}}
\def \tbR {\tilde{\bR}}
\def \tbU {\tilde{\bU}}
\def \tbE {\tilde{\bE}}
\def \tbF {\tilde{\bF}}

\def \tbOmega {\tilde{\bOmega}}
\def \tbGamma {\tilde{\bGamma}}
\def \tbtau {\tilde{\btau}}
\def \tbsigma {\tilde{\bsigma}}

\def \ta {\tilde{a}}
\def \tc {\tilde{c}}

\def \tp {\tilde{p}}
\def \tt {\tilde{t}}
\def \tx {\tilde{x}}
\def \ty {\tilde{y}}

\def \tF {\tilde{F}}
\def \tG {\tilde{G}}

\def \tU {\tilde{U}}
\def \tV {\tilde{V}}

\def \tGamma {\tilde{\Gamma}}
\def \tOmega {\tilde{\Omega}}

\def \tgrad {\tilde{\grad}}

\def \mF {\mathcal{F}}
\def \mI {\mathcal{I}}
\def \mM {\mathcal{M}}
\def \mR {\mathcal{R}}
\def \mV {\mathcal{V}}
\def \mO {\mathcal{O}}

\def \lp {\left(}
\def \rp {\right)}

\def \ls {\left[}
\def \rs {\right]}

\def \d {\text{d}}
\def \dr {\d r}

\def \tran {\mathsf{T}}

\newcommand{\nus}{Department of Mechanical Engineering, National University of Singapore, 117575, Singapore}

\title{\TITLE}

\author{Zhuoqun Xu~\orcid{0000-0002-3535-4402}}%
\affiliation{\nus}
\author{Lailai Zhu~\orcid{0000-0002-3443-0709}}%
\email{lailai\_zhu@nus.edu.sg}%
\affiliation{\nus}
\date{\today}%

\begin{abstract}
Machine learning and artificial intelligence have recently represented a popular paradigm for designing and optimizing robotic systems across various scales. Recent studies have showcased the innovative application of large language models (LLMs) in industrial control~\cite{song2023pre} and in directing legged walking robots~\cite{wang2023prompt}. In this study, we utilize an LLM, GPT-4, to train two prototypical microrobots for swimming in viscous fluids. Adopting a few-shot learning approach, we develop a minimal, unified prompt composed of only five sentences. The same concise prompt successfully guides two distinct articulated microrobots---the three-link swimmer and the three-sphere swimmer---in mastering their signature strokes. These strokes, initially conceptualized by physicists, are now effectively interpreted and applied by the LLM, enabling the microrobots to circumvent the physical constraints inherent to micro-locomotion. Remarkably, our LLM-based decision-making strategy substantially surpasses a traditional reinforcement learning method in terms of training speed. We discuss the nuanced aspects of prompt design, particularly emphasizing the reduction of monetary expenses of using GPT-4.
\end{abstract}

\maketitle
A fundamental characteristic of living organisms is their capacity for locomotory motion, encompassing a spectrum of  movements such as running, crawling, flying, slithering, and swimming.  Having evolved through the long process of natural selection and adaptation, these motility patterns observed in biological realm inspire the design of bionic articulated robots. They achieve locomotion by actuating their movable components similar to natural joints (or hinges and links), \eg walking legs, flapping wings, and undulating fins. 

A natural question arises: How can we actuate these joints concertedly to enhance the robot's locomotory performance, environmental adaptiveness, and perturbation resilience?  It corresponds to identifying the effective time sequence of robotic actions---forces (and torques) exerted on the moving elements or their translational (and rotational) velocities.

This question has be addressed by various means. For example, biological strategies from evolution can be mimicked with heuristic adjustment. Besides, diverse schemes in the frame of optimal control theory have been established for robotic movement~\cite{abdallah1991survey,wieber2016modeling,westervelt2018feedback}. Moreover, machine learning has expedited the development of more autonomous and intelligent robotic controllers. Notably, reinforcement learning (RL) as a popular machine learning technique has been applied to  walking~\cite{benbrahim1997biped, kimura2002reinforcement, haarnoja2018learning}, flying~\cite{xu2019learning,becker2020learning}, and swimming~\cite{gazzola2016learning,colabrese2017flow,verma2018efficient,schneider2019optimal,mirzakhanloo2020active, tsang2020self, qiu2020swimming, jiao2021learning, deng2021design, alageshan2020machine, muinos2021reinforcement,gerhard2021hunting,gunnarson2021learning, behrens2022smart,liu2022fishgym,zhu2022learning,wang2022learn,borra2022reinforcement,chen2022reinforcement,zou2022gait,nasiri2022reinforcement,zhu2022optimizing,qin2023reinforcement,el2023steering,mo2023chemotaxis,hu2023guided,lin2023emergence,10.1007/978-3-031-47258-9_20} robots at different scales. 

Yet, the recent surge in large language models (LLMs) or foundation models in general has spawned a new machine learning paradigm, in-context learning (ICL)~\cite{dong2022survey}, for decision-making and robotic manipulation.
Unlike RL or other machine learning techniques, ICL precludes updating neural network weights, instead learns to solve new tasks during the inference phase via receiving text prompts that incorporate exemplar task demonstrations. Owing to their extensive pre-training on diverse and huge datasets, LLMs exhibit the ICL capacity and thus can perform a variety of complex tasks via ICL including reasoning~\cite{wei2022chain}, logic deduction~\cite{suzgun2022challenging,creswell2022selection}, solving mathematical problems~\cite{lewkowycz2022solving}, 
sentiment analysis~\cite{mao2022biases}, language translation~\cite{agrawal2022context}, code generation~\cite{vaithilingam2022expectation}, medical diagnostics ~\cite{liu2023deid}, and so on.
More recently, LLM has been utilized for industrial and robotic control. In particular, GPT-4, an LLM pre-rained on internet-scale datasets, has been successfully utilized for controlling HVAC~\cite{song2023pre} and for enabling terrestrial robots to walk~\cite{wang2023prompt} in simulated environments. Importantly, Ref.~\cite{song2023pre} reveals that this LLM approach attains a performance comparable to that of RL, but requiring fewer samples and incurring lower technical debt.

\begin{figure*}[tbh!]
\centering
\includegraphics[width=0.8\linewidth]{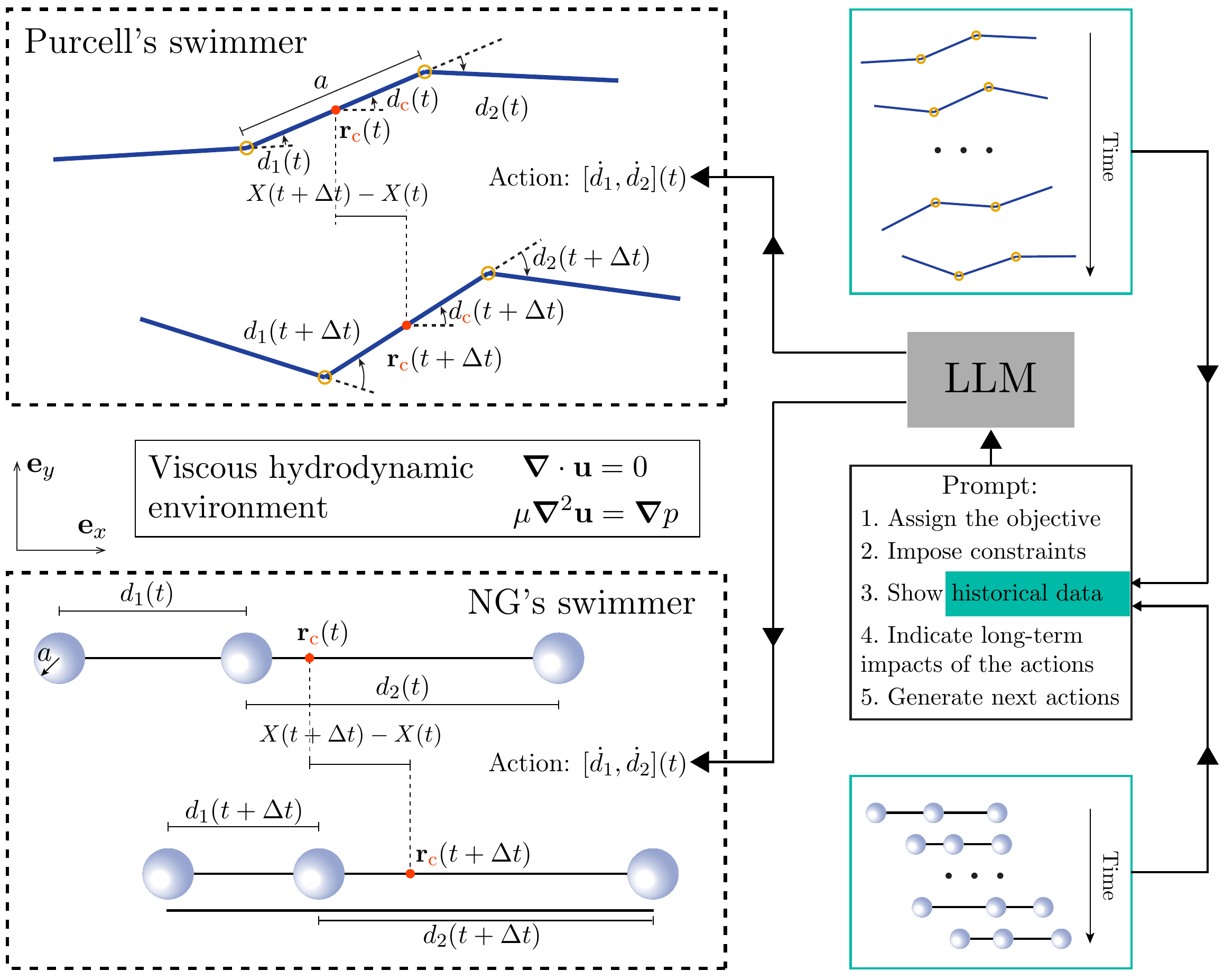}
\caption{
Diagrammatic representation of utilizing an LLM, GPT-4 adopted here, for microrobotic locomotion. We have developed a minimal, unified minimal prompt capable of instructing two
model microswimmers---Purcell's swimmer~\cite{purcell1977life} and NG' swimmer~\cite{najafi2004simple}---to propel through highly viscous fluids. These microswimmers' movements are subject to viscous hydrodynamics governed by the Stokes equations: $\grad \cdot \bu = 0$ and $\mu \grad^2 \bu = \grad p$, where $\bu$ and $p$ denote the velocity and pressure fields, respectively, with $\mu$ indicating the fluid's dynamic viscosity. 
Comprising only five sentences, the prompt effectively orchestrates the interaction between the swimmer and the LLM, directing the former to swim along the horizontal axis ($\be_x$ or $-\be_x$) with maximal speed. The swimmer's displacement is $X = \brc \cdot \be_x$, where $\brc$ denotes its geometric centroid.
}
\label{fig:1}
\end{figure*}

In this work, we explore the ability of LLMs
to deduce physical principles underlying robotic locomotion and subsequently leverage the understanding to design 
locomotory gaits.
Our specific focus is on microrobotic swimming at a vanishing Reynolds ($\Re$) number, a scenario physically constrained by the dominance of viscous fluid forces over inertia forces. This configuration enables examining whether LLMs can perceive Purcell's ``scallop theorem'', introduced by Edward Purcell in his seminal 1977 lecture ``Life at low Reynolds number''~\cite{purcell1977life}. The theorem posits that in the inertialess (Stokesian) limit, \ie $\Re=0$, reciprocal motions or actuations, exemplified by a scallop's opening and closing, cannot generate net movement. Consequently, a one-hinged microswimmer fails in effective propulsion in viscous fluids, because its single degree of freedom (DOF) leads to inherently reciprocal movement. To circumvent this physical constraint, he then proposed a two-hinged `simplest animal'. Later known as Purcell's three-link swimmer (see Fig~\ref{fig:1} left upper panel), the minimal model can effectively propel through non-reciprocal motions, enabled by the addition of more than one DOF.

Here, we explore the LLM, GPT-4, 
to navigate this Purcell's three-link swimmer and another prototypical model---Najafi-Golestanian (NG)’s three-sphere swimmer~\cite{najafi2004simple} in inertialess viscous fluids. These 
swimmers
communicate with the LLM by responding to its generated actions and sending a few task demonstrations back through text prompts. We design a unified minimal prompt framework to coordinate the interaction between GPT-4 and both swimmers. Remarkably, despite its training lacking physical data, GPT-4 enables the microswimmers to learn efficient strokes, thereby overcoming the low-Reynolds-number physical constraints.

\section*{\secone}
The Purcell's swimmer comprises three slender cylindrical links of length $a$ and radius $b$, connected in sequence by two planar hinges. These hinges allow the links to rotate relative to their neighbor(s). Propulsion is attainable by varying over time ($t$) the relative angles, $\dofone (t)$ and $\doftwo (t)$,
between every two adjacent links. In contrast, NG's swimmer constitutes three spheres with radius $a$, aligned colinearly along their common axis along $\be_x$. The spheres are linked sequentially by two extensible links, with lengths $\dofone (t)$ and $\doftwo (t)$, respectively. Time-varying the two DOFs enables this swimmer to translate in the $\be_x$ direction. 
Notably, we have intentionally adopted the same notation $a$ as the characteristic length of both  models. 
The same intention applies to their DOFs, $\dofone$ and $\doftwo$.

We characterize the locomotion of both swimmers by the velocities of their centroid coordinated at $\brc$. Namely, NG's swimmer translates horizontally at a velocity of $ \dotbrc \be_x$,
with the overdot signifying a time derivative.
In contrast, the Purcell's swimmer exhibits two-dimensional motion, reflected by the translational velocity $\dotbrc$ of its middle link's centroid and the link's rotational velocity $\dotdofc \bez$. Below, we will briefly describe how to calculate the swimming velocity of Purcell's swimmer. The corresponding calculation for NG's swimmer is given in \mat.

\begin{figure*}[tbh!]
\centering
\includegraphics[width=1\linewidth]{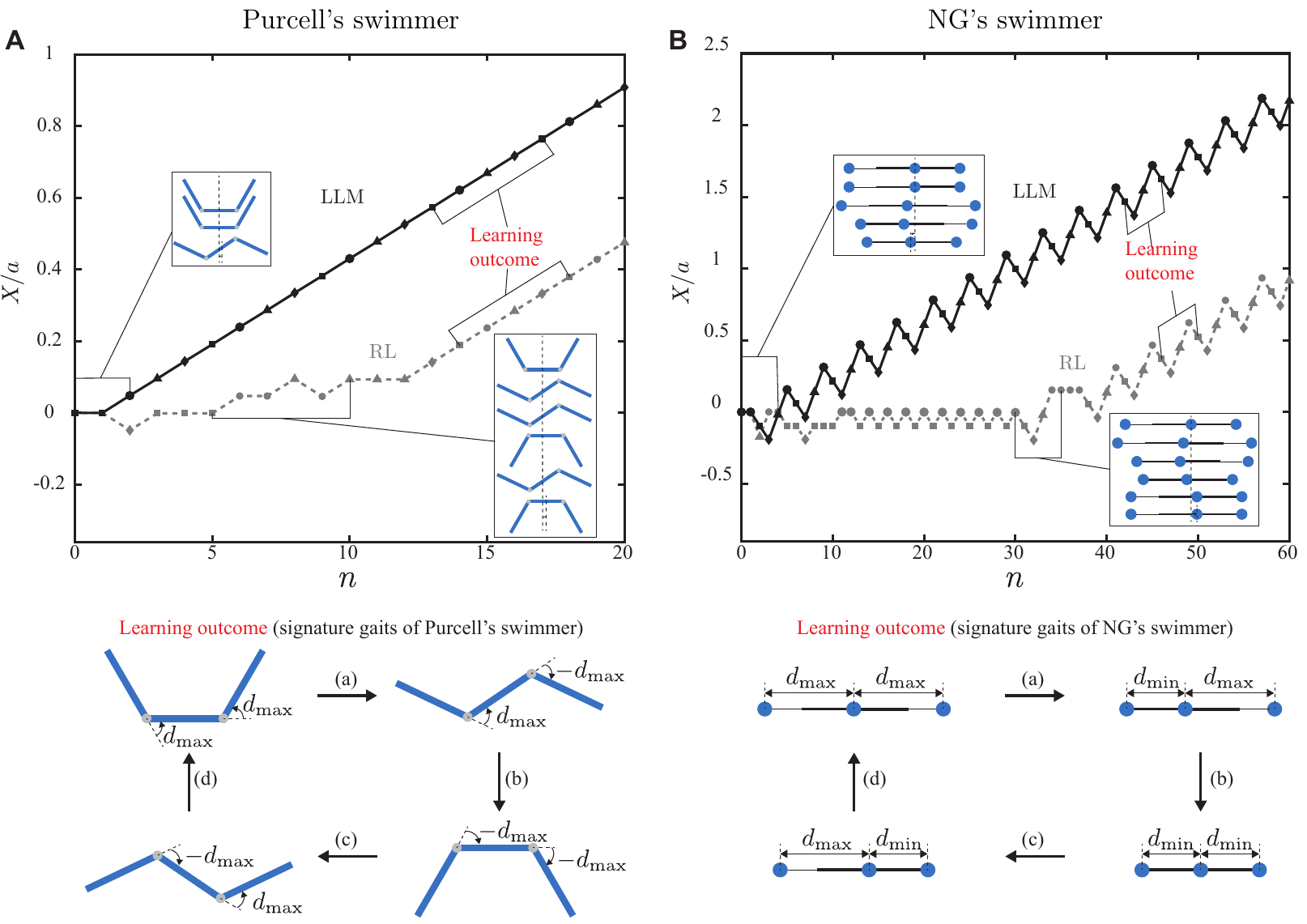}
\caption{A, displacement $X/a$ of Purcell's swimmer versus its execution step $n$, which is trained by the LLM (solid line) and $Q$-learning-based RL (dashed line).  The lower panel demonstrates the cycle of signature gaits learned by the swimmer. B, same as A, but for NG's swimmer.
}
\label{fig:gait}
\end{figure*}
In this study of Purcell's swimmer, we allow only one angular DOF, either $\dofone$ or $\doftwo$, to vary at a time. The selected DOF is restricted to a discrete set of angles, $\left[-\dofmax, \dofmax \right]$. Hence, the swimming action as the DOF' rate of change (ROC) is also discrete and drawn from $\left[-2\dofmax/T, 0, 2\dofmax/T \right]$, where $T$ represents the gear-changing time. 

To map the swimmer' stroke to its kinematics, we assume that its links  possess a small slenderness $\epsilon = b/a \ll 1$. This allows for using the local resistive force theory to account for the swimming hydrodynamics.
For a link parameterized by the arc-length $s\in[0,a]$,  the hydrodynamic force per unit length $\bf$ is:
\begin{align}
    \bf(s)=  - \left[ \xi_{\parallel}\bt\bt + \xi_{\perp}\lp \bI -\bt\bt \rp \right] \cdot \dot{\br}(s),
\end{align}
where 
$\xi_{\parallel}$ and $\xi_{\perp}$ 
represent the resistive force coefficients, and $\bt$ and $\br(s)$ denote the link's tangent vector and local coordinates, respectively. Here, $\bI$ is the identity tensor. We consider the limit $\epsilon \rightarrow 0$ here, leading to $\xi_{\perp}/\xi_{\parallel}\rightarrow 2$.
In the inertialess flow, the swimmer's translational and rotational velocities can be derived using the constraints of net zero hydrodynamic force and torque summed over all links, respectively. For an individual link, the force is $\int_0^a \bf (s) \d s$, and the torque about the centroid is $\int_0^a \left[ \br (s) - \brc \right] \times \bf(s) \d s$. Applying the constraints, we derive the swimming velocities analytically using \textit{Mathematica}, with their formula given in \mat.

We build the physical environment in a dimensionless framework. By choosing $2\dofmax/T$ as the characteristic angular velocity,  the dimensionless ROCs fall within the set of discrete integers, $[-1,0,1]$. This integer nature significantly aids ICL via LLMs, as these models commonly have difficulty in accurately interpreting floating numbers~\cite{mirchandani2023large}.

\section*{\sectwo}

We employ the few-shot prompting method with standard GPT-4, instructing it to directly generate actions for a microswimmer without task-specific fine-tuning. The swimmer subsequently executes the received action in the physical environment, with the latest few historical snapshots stored in a textual buffer. This buffer is then provided to GPT-4, aiding its subsequent decision-making.

We have designed a unified prompt that can guide both swimmers effectively, with a particular focus on minimizing the text length and associated monetary cost. Our prompt comprises five functional sections, each containing a single sentence, thereby highlighting the minimal textual elements required.  Firstly, it sets the objective: to determine the ROC sequence in both DOFs, targeting the swimmer's fastest long-term movement in a prescribed direction. 
The second sentence define constraints on the DOFs, \eg $\dofone \in \left[-\dofmax,\dofmax \right]$. 
Third, we demonstrate the latest $\ns$ historical records of the interactions between the LLM and the physical environment, leaving $\ns$ as a tunable parameter.
The following prompt alerts the LLM to the potential for long-term impacts of the action. 
Finally, the last sentence instructs the LLM to generate the next action constrained within a specified group.

\subsection{History clearing scheme}
Occasionally, we observe that the swimmer cannot develop an effective strategy, becoming ensnared in a cycle of erroneous strokes. This trapping failure may be attributed to the limitations inherent in the few-shot prompting method: the LLM struggles to suggest appropriate actions based solely on a brief history of unsuccessful attempts and their adverse outcomes.

To circumvent this issue, we propose and implement a history clearing scheme. Specifically, when the swimmer's displacement within the last few (this number is predetermined) steps falls below a threshold, such as $0.01$, we erase and subsequently rebuild the entire record of prior few-shot demonstrations. In practice, this simple approach has proven to be successful in preventing the aforementioned entrapment.

\subsection{Avoiding floating-point or negative numbers}
LLMs are known to struggle with comprehending floating-point or negative numbers~\cite{wang2023prompt}. To address this issue, we implement a transformation to the swimmer's displacement $X/a$. Its raw value  produced by the theoretical model is a floating number truncated to three decimal places, a step taken to curtail monetary expenses. 
By predefining a lower boundary, $X_{\text{min}}/a$, for $X/a$, we apply the following transformation $\left(X/a\right)_{\rm LLM} = \left(X/a - X_{\text{min}}/a\right) \times 1000$, upon which the LLM receives a transformed value $\left(X/a\right)_{\rm LLM}$ as a positive integer.

\subsection{Saving money by using aliases extensively}
Throughout our study, we have emphasized  saving the
monetary costs of using GPT-4, primarily by minimizing the text length. 
We find that a simple approach---extensively employing aliases---significantly aids in minimization. For example, we use the alias `DOF' in place of `degree of freedom', and `ROC' for `rate of change'. GPT-4 seems to fully comprehend such abbreviations, thereby enabling the feasibility of this cheap strategy.

\subsection{Temperature (randomness) within GPT-4}
In the GPT-4 architecture, the `temperature' parameter indicative of the level of randomness, can be adjusted within the range $[0,1]$. This adjustment allows for a balance between consistent output and creative exploration by LLM. Throughout this study, we have set the temperature parameter to zero,  focusing on deterministic evaluations.

\begin{figure*}[thb!]
\centering
\includegraphics[width=1\linewidth]{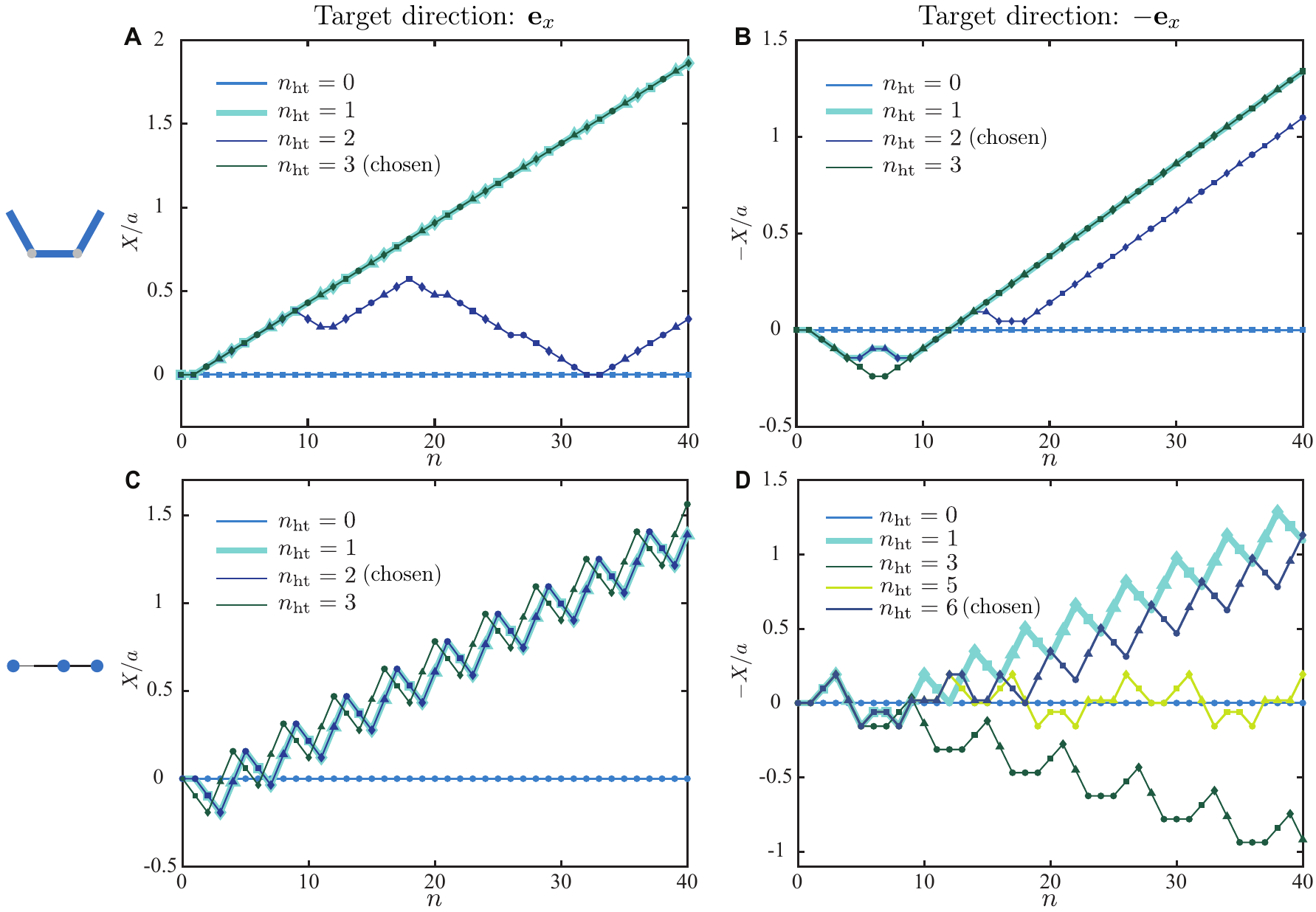}
\caption{Influence of the length $\ns$ of historical records on the swimmers' learning performance---the swimmer's displacement in the target direction, $X/a$ (positive $\be_x$, left column) or $-X/a$ (negative $\be_x$, right column) versus the execution step $n$.
The upper and lower rows correspond to Purcell's swimmer and NG's swimmer, respectively. 
}
\label{fig:hist}
\end{figure*}

\section*{Results and observations}
In this section, we demonstrate using GPT-4 to 
guide the swimming microrobots. In general, the performance
depends on their initial configuration, the number of historical demonstrations $\ns$, and the target direction (positive or negative $\be_x$ here).

\subsection*{\secthree}
As shown in Fig.~\ref{fig:gait}, GPT-4 effectively directs both model microrobots to swim consistently in a specific direction---positive $\be_x$ here. Remarkably, using the LLM as the decision maker, both robots take only one execution step to acquire the well-known signature cycle of strokes (see the lower panels of Fig.~\ref{fig:gait}). Such strokes, non-reciprocal in time, have originally been perceived by physicists~\cite{purcell1977life,najafi2004simple} to overcome the low-Reynolds-number constraints. Moreover, the LLM-based control dramatically outperforms
the RL approach in terms of the acquisition speed. Trained by a $Q$-learning scheme, the RL agents for Purcell's and NG's swimmers require, respectively, $\approx 12$ and $\approx 40$ steps before learning these strokes. Though Fig.~\ref{fig:gait} illustrates the training process for one specific initial configuration of both swimmers, we assert that the same prompt effectively function for all the other three initial configurations.

In our LLM prompt, the number of historical demonstration, $\ns$, is an important parameter influencing the learning performance. A larger $\ns$ facilitates ICL, yet increasing the information exchanged between the prompt and GPT-4, consequently raising the monetary cost. As illustrated in Fig.~\ref{fig:hist}A \& B, Purcell's swimmer, with $\ns=1$, successfully learns effective gaits when directed to swim in both the positive and negative $\be_x$ directions. However, we refrain from selecting $\ns=1$ for other parts of this study. This decision is based on observations that the learning performance degenerates when $\ns$ is increased to $2$ in both directions, suggesting that the success at $\ns=1$ is mostly attributed to a fortuitous coincidence rather than a systematic pattern. Likewise, as shown in Fig.~\ref{fig:hist}C \& D,
we opt for $\ns=3$ and $\ns=6$ for NG's swimmer when navigating in the $\be_x$ and $-\be_x$ directions, respectively. To be rigorous, we do not designate these selected $\ns$ as the minimum necessary number of historical shots.

We further examine the robustness of this LLM-based decision making process under thermal noise in the swimming environment. Specifically, we randomly perturb the swimmer's incremental displacement $\Delta X$ following each execution step, yielding a noisy movement $\Delta X \lp 1 + \zeta Y \rp$. Here, $\zeta$ indicates the noise level, and $Y$ is a random number uniformly distributed in $[-1,1]$. 

\begin{figure}[tbh!]
\centering
\includegraphics[width=1\linewidth]{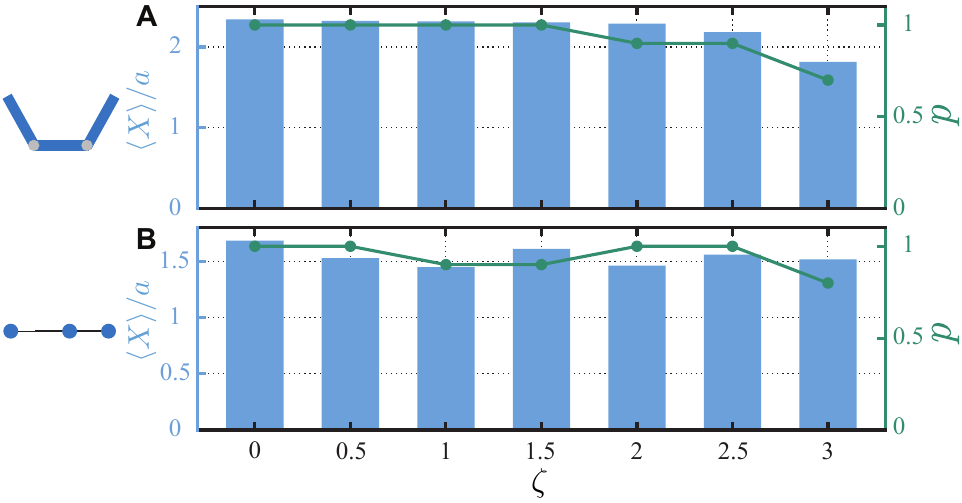}
\caption{Average displacement $\mX/a$ and overall success rate $p$ for Purcell's swimmer (A) and NG's swimmer (B) guided along the $\be_x$ direction, under varying noise levels $\zeta$. For each level, $10$ individual runs are conducted to obtain the statistics.
}
\label{fig:noise}
\end{figure}

For each noise level, we perform $10$ tasks, evaluating the average displacement $\mX/a$ and the overall success rate $p$. The number of tasks, $10$, is selected to  balance statistical significance with monetary expenses.
A task is considered successful if the swimmer learns to execute three consecutive cycles of its signature gaits within $50$ steps, without subsequent failures. Fig.~\ref{fig:noise} illustrates how the noise $\zeta$ affects the swimmer's learning performance. With increasing $\zeta$, we generally observe a decrease in both $\mX/a$ and $p$, although the decline in performance is relatively modest. Notably, the success rate remains around $p \approx 0.75$ at the highest noise level of $\zeta=3$, indicating the overall resilience of this LLM-based approach.

\subsection*{\secfive}

\begin{figure}[tbh!]
\centering
\includegraphics[width=1\linewidth]{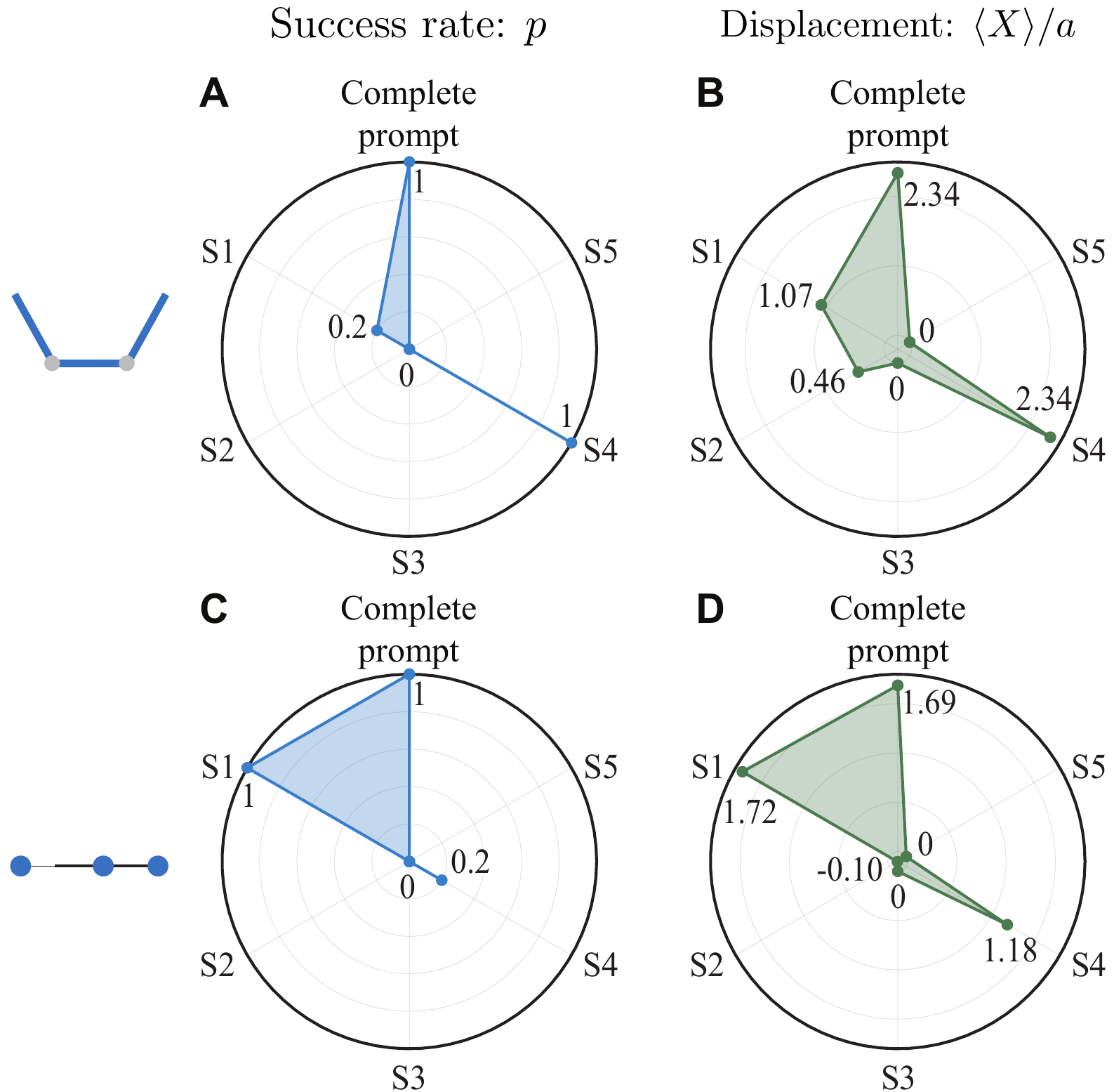}
\caption{Criticality of the five sentences, labeled S1 to S5, in the prompt. Left column: how the success rate $p$ of Purcell's swimmer (A) and NG's swimmer (C) guided in the $\be_x$ direction is affected when a single sentence is removed from the prompt, in comparison to the results obtained from the complete prompt (12 o'clock in the pie chart). The right column mirrors the left, but instead evaluates the impact on the average displacement $\mX/a$, rather than the success rate $p$.
}
\label{fig:del_sen}
\end{figure}

We note that our prompt is significantly simpler and shorter than those in related works~\cite{song2023pre,wang2023prompt}, mostly owing to the simplicity of our physical environment. The major simplification we have made is summarized below. First, unlike these works, we do not use a system log to provide an overview of the task. It turns out to be unnecessary and is thus removed for reducing monetary cost.
Secondly, we do not employ a trained controller as in Ref.~\cite{wang2023prompt} to collect the initial batch of historical demonstrations fed to the LLM. Neither do we incorporate any expert demonstrations dataset as in Ref.~\cite{song2023pre} that facilitates the LLM to learn the working principles underlying expert policies.

Furthermore, we have investigated whether the designed prompt is minimal by assessing the indispensability of each of the five sentences, labeled S1 to S5. To evaluate the necessity of an individual sentence, we conducted $10$ runs with the respective sentence omitted from the prompt. We then analyzed the overall success rate $p$ and average displacement $\mX/a$ of both swimmers, as shown in Fig.~\ref{fig:del_sen}. Our comparative analysis reveals that sentences S2, S3, and S5 are critically indispensable; the absence of any of these results in failure of all runs. Conversely, sentences S1 and S4 demonstrate weaker indispensability, with occasional task success despite their removal.

\section*{\seccon}\label{sec:con}

In this work, we have demonstrated the hitherto unreported usage of LLMs for effectuating low-Reynolds-number propulsion of model microswimmers.
Employing a few-shot prompting approach, we have designed a five-sentence prompt for two minimal swimmers: Purcell's three-link swimmer and NG's three-sphere swimmer. This unified prompt enables GPT-4 to generate through ICL, the optimal stroke patterns for both swimmers, despite their distinct propulsion mechanisms.

Based on the current successful demonstration of using LLMs for microrobotic manipulation, we identify several potentially promising research directions: 1) transitioning from a discrete to a continuous action space; 2) steering microrobots to navigate through complex environments encompassing non-Newtonian fluid rheology, complex confinements, and obstacles; 3) facilitating cooperative swimming behaviors among microrobots.

We admit that certain hyperparameters in this preliminary study were selected arbitrarily. For instance, we set the temperature parameter of GPT-4 to zero. In our forthcoming research, we aim to examine the influence of this temperature setting and explore whether increasing it could potentially improve the exploratory capabilities of the LLM.

\newpage
\section*{Materials and Methods}
\subsection*{Hydrodynamics of Purcell's Swimmer}

For Purcell's swimmer, the middle link  moves with a translational velocity $\dotbrc = u \bex + v \bey$ and rotates with an angular velocity $\dotdofc \bez$ about its center. The two velocities as a function of actions $\lp \dotdofone,\dotdoftwo \rp$ are calculated by \textit{Mathematica} and are given by \eqnrefS{eq:SI_u}, \eqnrefS{eq:SI_v}, and \eqnrefS{eq:SI_rot}. For brevity, 
all the variables in these equations are dimensionless, where we have chosen the rod length $a$ as the characteristic length, $T/2\dofmax$ the characteristic time, and $2\dofmax/T$ the characteristic angular velocity.

\begin{figure*}[ht]
\centering
\begin{equation}\label{eq:SI_u}
\begin{split}
    u = \frac{1}{2 \Delta} \left(\dotdofone \left(262 \sin \left(\dofc -\dofone\right)-26 \sin \left(\dofone+\dofc \right)-2 \sin \left(2 \dofone+\dofc \right)+18 \sin \left(\dofc -2 \dofone\right)-16 \sin \left(\dofc -\doftwo\right)+78 \sin \left(-\dofone-\doftwo+\dofc \right) \right. \right. \quad &\\   -18 \sin \left(\dofone-\doftwo+\dofc \right)+8 \sin \left(-2 \dofone-\doftwo+\dofc \right)+104 \sin \left(\doftwo+\dofc \right)+126 \sin \left(-\dofone+\doftwo+\dofc \right)+30 \sin \left(\dofone+\doftwo+\dofc \right)&\\  +24 \sin \left(2 \doftwo+\dofc \right)+21 \sin \left(-\dofone+2 \doftwo+\dofc \right)+\sin \left(\dofone+2 \doftwo+\dofc \right)-2 \sin \left(2 \dofone+2 \doftwo+\dofc \right)-4 \sin \left(\dofc -2 \doftwo\right)&\\  \left.+9 \sin \left(-\dofone-2 \doftwo+\dofc \right)-3 \sin \left(\dofone-2 \doftwo+\dofc \right)+2 \sin \left(-2 \dofone-2 \doftwo+\dofc \right)+36 \sin (\dofc )\right)+\dotdoftwo \left(104 \sin \left(\dofc -\dofone\right)-16 \sin \left(\dofone+\dofc \right) \right.&\\ -4 \sin \left(2 \dofone+\dofc \right)+24 \sin \left(\dofc -2 \dofone\right)-26 \sin \left(\dofc -\doftwo\right)+30 \sin \left(-\dofone-\doftwo+\dofc \right)-18 \sin \left(\dofone-\doftwo+\dofc \right)-3 \sin \left(2 \dofone-\doftwo+\dofc \right) &\\   +\sin \left(-2 \dofone-\doftwo+\dofc \right)+262 \sin \left(\doftwo+\dofc \right)+126 \sin \left(-\dofone+\doftwo+\dofc \right)+78 \sin \left(\dofone+\doftwo+\dofc \right)+9 \sin \left(2 \dofone+\doftwo+\dofc \right)&\\  +21 \sin \left(-2 \dofone+\doftwo+\dofc \right)+18 \sin \left(2 \doftwo+\dofc \right)+8 \sin \left(\dofone+2 \doftwo+\dofc \right)+2 \sin \left(2 \dofone+2 \doftwo+\dofc \right)-2 \sin \left(\dofc -2 \doftwo\right)&\\ \left. \left. -2 \sin \left(-2 \dofone-2 \doftwo+\dofc \right)+36 \sin (\dofc )\right)\right).
\end{split}
\end{equation}
\end{figure*}

\begin{figure*}[ht]
\centering
\begin{equation}\label{eq:SI_v}
\begin{split}
    v = -\frac{1}{2 \Delta} \left(\dotdofone \left(262 \cos \left(\dofc -\dofone\right)-26 \cos \left(\dofone+\dofc \right)-2 \cos \left(2 \dofone+\dofc \right)+18 \cos \left(\dofc -2 \dofone\right)-16 \cos \left(\dofc -\doftwo\right)+78 \cos \left(-\dofone-\doftwo+\dofc \right)\right. \right. &\\ -18 \cos \left(\dofone-\doftwo+\dofc \right)+8 \cos \left(-2 \dofone-\doftwo+\dofc \right)+104 \cos \left(\doftwo+\dofc \right)+126 \cos \left(-\dofone+\doftwo+\dofc \right)+30 \cos \left(\dofone+\doftwo+\dofc \right)&\\ +24 \cos \left(2 \doftwo+\dofc \right)+21 \cos \left(-\dofone+2 \doftwo+\dofc \right)+\cos \left(\dofone+2 \doftwo+\dofc \right)-2 \cos \left(2 \dofone+2 \doftwo+\dofc \right)-4 \cos \left(\dofc -2 \doftwo\right)&\\  \left. +9 \cos \left(-\dofone-2 \doftwo+\dofc \right)-3 \cos \left(\dofone-2 \doftwo+\dofc \right)+2 \cos \left(-2 \dofone-2 \doftwo+\dofc \right)+36 \cos (\dofc )\right)+\dotdoftwo \left(104 \cos \left(\dofc -\dofone\right)-16 \cos \left(\dofone+\dofc \right)\right. &\\ -4 \cos \left(2 \dofone+\dofc \right)+24 \cos \left(\dofc -2 \dofone\right)-26 \cos \left(\dofc -\doftwo\right)+30 \cos \left(-\dofone-\doftwo+\dofc \right)-18 \cos \left(\dofone-\doftwo+\dofc \right)-3 \cos \left(2 \dofone-\doftwo+\dofc \right)&\\ +\cos \left(-2 \dofone-\doftwo+\dofc \right)+262 \cos \left(\doftwo+\dofc \right)+126 \cos \left(-\dofone+\doftwo+\dofc \right)+78 \cos \left(\dofone+\doftwo+\dofc \right)+9 \cos \left(2 \dofone+\doftwo+\dofc \right)&\\ +21 \cos \left(-2 \dofone+\doftwo+\dofc \right)+18 \cos \left(2 \doftwo+\dofc \right)+8 \cos \left(\dofone+2 \doftwo+\dofc \right)+2 \cos \left(2 \dofone+2 \doftwo+\dofc \right)-2 \cos \left(\dofc -2 \doftwo\right)&\\ \left. \left.-2 \cos \left(-2 \dofone-2 \doftwo+\dofc \right)+36 \cos (\dofc )\right)\right).
\end{split}
\end{equation}
\end{figure*}

\begin{figure*}[ht]
\centering
\begin{equation}\label{eq:SI_rot}
\begin{split}
    \dotdofc = \frac{2}{\Delta} \left(\dotdofone \left(102 \cos \left(\dofone\right)+2 \cos \left(2 \dofone\right)-6 \cos \left(\dofone-\doftwo\right)-4 \cos \left(2 \doftwo\right)+42 \cos \left(\dofone+\doftwo\right)+2 \cos \left(2 \left(\dofone+\doftwo\right)\right)+9 \cos \left(\dofone+2 \doftwo\right) \right. \right. \phantom{xxx} &\\ \left. -3 \cos \left(\dofone-2 \doftwo\right)+108\right)+\dotdoftwo \left(4 \cos \left(2 \dofone\right)+6 \cos \left(\dofone-\doftwo\right)+3 \cos \left(2 \dofone-\doftwo\right)-102 \cos \left(\doftwo\right)-2 \cos \left(2 \doftwo\right)-42 \cos \left(\dofone+\doftwo\right)\right. &\\ \left. \left.-2 \cos \left(2 \left(\dofone+\doftwo\right)\right)-9 \cos \left(2 \dofone+\doftwo\right)-108\right)\right).
\end{split}
\end{equation}
\end{figure*}

\begin{figure*}[ht]
\centering
\begin{equation}\label{eq:SI_Delta}
\begin{split}
    \Delta = 3 \left(136 \cos \left(\dofone\right)+14 \cos \left(2 \dofone\right)-8 \cos \left(\dofone-\doftwo\right)-\cos \left(2 \left(\dofone-\doftwo\right)\right)-4 \cos \left(2 \dofone-\doftwo\right)+14 \cos \left(2 \doftwo\right) +56 \cos \left(\dofone+\doftwo\right)\right. \phantom{XXXX} &\\  \left.+136 \cos \left(\doftwo\right)+3 \cos \left(2 \left(\dofone+\doftwo\right)\right)+12 \cos \left(2 \dofone+\doftwo\right)+12 \cos \left(\dofone+2 \doftwo\right)-4 \cos \left(\dofone-2 \doftwo\right)+282\right).
\end{split}
\end{equation}
\end{figure*}

\subsection*{Hydrodynamics of NG's Swimmer}
For NG's swimmer, we allow only one DOF, either $\dofone$ or $\doftwo$, to vary at a time. The selected DOF is restricted to a discrete set of lengths, $\left[\dofmin, \dofmax \right]$. Hence, the action is also discrete and drawn from $\left[-(\dofmax - \dofmin)/T, 0, (\dofmax - \dofmin)/T \right]$, where $T$ is the time a rod takes to change its length. Following Ref.~\cite{golestanian2008analytic}, the dimensionless velocity of the swimmer reads:
\begin{equation}
\begin{split}
    \dotbrc = \frac{1}{6}\left[\left( \frac{\dotdoftwo - \dotdofone}{\dofone + \doftwo}\right) + 2\left(\frac{\dotdofone}{\doftwo} - \frac{\dotdoftwo}{\dofone} \right)\right],
\end{split}
\end{equation}
where we have selected $a$ as the characteristic length, $\lp\dofmax-\dofmin\rp/T$ as the characteristic velocity, and $aT/\lp\dofmax-\dofmin\rp$ as the characteristic time for the non-dimensionalization.

\subsection*{Data Availability}
The datasets used and/or analysed during the current study available from the corresponding author on reasonable request.

\subsection*{Code Availability}
The GPT-4 prompt will be open-sourced upon the acceptance of this manuscript.


\clearpage

\end{document}